\documentclass[lettersize,journal]{IEEEtran}
\usepackage{amsmath,amsfonts}
\usepackage{algorithmic}
\usepackage{array}
\usepackage[caption=false,font=normalsize,labelfont=sf,textfont=sf]{subfig}
\usepackage{textcomp}
\usepackage{stfloats}
\usepackage{url}
\usepackage{verbatim}
\usepackage{multirow}
\usepackage{graphicx}
\usepackage{cite}
\usepackage{soul}
\usepackage{color, xcolor} 
\usepackage{booktabs}
\usepackage{pdflscape}
\usepackage{amsmath}
\usepackage{array} 
\usepackage{multirow}
\def\BibTeX{{\rm B\kern-.05em{\sc i\kern-.025em b}\kern-.08em
    T\kern-.1667em\lower.7ex\hbox{E}\kern-.125emX}}
\usepackage{balance}
\begin{document}
\title{Reference-Free Enhancement of Forward-Looking Sonar Images: Bridging Cross-Modal Degradation Gaps Through Deformable Wavelet Scattering Transform and Multi-Frame Fusion}
\author{
\IEEEauthorblockN{Zhisheng Zhang$^{\dagger}$, Peng Zhang$^{\dagger}$, Fengxiang Wang, Liangli Ma$^{*}$, Fuchun Sun$^{*}$, \IEEEmembership{Fellow, IEEE}}

\thanks{
Zhisheng Zhang is with the College of Electronic Engineering, Naval University of Engineering, Wuhan 430033, China. E-mail: d23181004@nue.edu.cn.}
\thanks{
Peng Zhang is with the College of Meteorology and Oceanography, National University of Defense Technology, Changsha 410073, China. E-mail: zhangpeng23a@nudt.edu.cn.}
\thanks{
Fengxiang Wang is with the College of Computer Science and Technology, National University of Defense Technology, Changsha 410073, China. E-mail: wfx23@nudt.edu.cn.}
\thanks{
Liangli Ma is with the College of Electronic Engineering, Naval University of Engineering, Wuhan 430033, China. E-mail: startmll@163.com.}
\thanks{
Fuchun Sun is with the Department of Computer Science and Technology, Tsinghua University, Beijing 100084, China. E-mail: fcsun@tsinghua.edu.cn.}

\thanks{*Corresponding authors: Fuchun Sun (fcsun@tsinghua.edu.cn) and Liangli Ma (startmll@163.com).}
 \thanks{$^{\dagger }$ Equal contribution}
 }

\maketitle

\begin{abstract}
Enhancing forward-looking sonar images is crucial for effective underwater target detection. Existing deep learning approaches typically depend on supervised training using simulated data, but their performance and generalization are limited by the scarcity of real-world paired datasets. Although recent reference-free methods have mitigated reliance on paired data, they overlook the inherent cross-modal degradation gap between sonar and remote sensing images. Direct application of pretrained remote-sensing models to sonar data frequently yields overly smoothed images with lost details and inadequate brightness. To overcome this challenge, we propose a novel Deformable Wavelet Scattering Transform (WST) Feature Bridge, which adaptively projects sonar images into a multi-scale, translation-invariant feature domain, effectively narrowing the cross-modal degradation gap. Additionally, we develop a reference-free multi-frame fusion strategy that exploits complementary inter-frame information to suppress speckle noise and enhance target visibility without external ground-truth. Experiments conducted on three sonar datasets demonstrate that the proposed framework substantially surpasses existing reference-free methods in terms of noise suppression, edge preservation, and brightness enhancement, highlighting its practical effectiveness for underwater detection scenarios.
\end{abstract}

\begin{IEEEkeywords}
Forward-Looking Sonar, reference-free, Multi-Frame Fusion, Cross-Modal Degradation Gap.
\end{IEEEkeywords}

\section{Introduction} 
\IEEEPARstart{F}{orward-looking} sonar (FLS) provides unique underwater imaging capabilities, becoming essential in underwater target detection, environmental perception, and autonomous navigation tasks~\cite{r1,r2,r3,r4,sr1}. However, complex underwater environments significantly degrade sonar images due to multipath scattering, acoustic absorption by target materials, and reflections from environmental boundaries. Such effects typically introduce severe speckle noise, brightness inconsistency, and various imaging artifacts~\cite{r5,r6,r7}, hindering accurate target detection and reliable 3D reconstruction, thereby limiting practical deployment of FLS in real-world scenarios~\cite{r8,r9,r10}.
\begin{figure}[!t]
    \centering
    \includegraphics[width=0.45\textwidth]{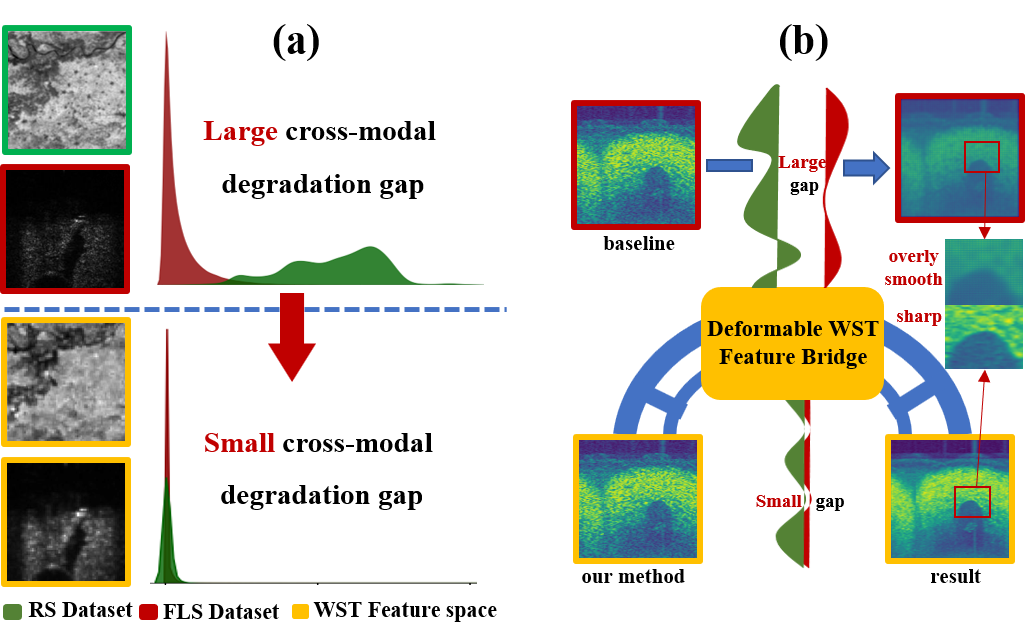}	
    \caption{(a) Bridging the Cross-Modal Gap. The degradation gap between multispectral remote sensing (RS) and forward-looking sonar (FLS) images is effectively narrowed by mapping both modalities into a unified feature space via our Deformable WST Feature Bridge; (b) Enhanced Results with Our Method. Compared to the overly smoothed baseline, our method preserves clear target contours and distinct foreground-background boundaries.}
    \label{fig1}
\end{figure}

Current sonar image enhancement methods generally fall into two categories~\cite{r11,r12,r13}: traditional filtering-based techniques and supervised deep-learning approaches. Traditional methods, including filtering, contrast adjustment, and grayscale transformations~\cite{tr1,tr2,tr3}, often struggle to restore severely degraded sonar images effectively. Supervised deep-learning approaches~\cite{sup1,sup2,sup3,sup4}, while capable of achieving better enhancement, heavily depend on large-scale, high-quality paired datasets. Collecting such datasets in realistic underwater environments is notoriously challenging, which severely restricts their generalizability~\cite{simu1,simu2}. Recent reference-free enhancement methods~\cite{unsup1,unsup2} circumvent this dependence but have yet to address the intrinsic cross-modal degradation differences between forward-looking sonar and multispectral remote sensing images~\cite{fx1,fx2,sr2}. Although both modalities exhibit speckle noise, remote sensing images suffer from blur and low contrast due to uneven illumination and atmospheric attenuation under a pinhole camera model~\cite{fx3,fx4}, whereas sonar imaging is fundamentally distance-based, resulting in multipath scattering-induced brightness variations and weaker contrast~\cite{r1,r2}. Consequently, directly applying pretrained remote-sensing models to sonar enhancement frequently produces overly smoothed outputs with loss of essential details (Fig.~1(a), (b)).

To address these limitations, we propose a novel reference-free sonar enhancement pipeline that integrates a Deformable Wavelet Scattering Transform (WST) Feature Bridge and a multi-frame fusion network into an end-to-end framework. Specifically, our Deformable WST Feature Bridge employs learnable perturbations on wavelet scales and orientations—carefully selected via ablation among multiple handcrafted features~\cite{fe1,fe2,fe3,tr1,SAR}—to adaptively map sonar images into a robust, translation-invariant feature domain, significantly narrowing the cross-modal degradation gap. Subsequently, our multi-frame fusion network leverages complementary information from consecutive sonar frames, naturally suppressing speckle noise and enhancing target-region brightness without relying on external high-quality ground truth. Unlike previous methods, our unified approach simultaneously optimizes adaptive feature extraction and multi-frame fusion, resulting in sharper contours, richer details, and enhanced brightness in sonar imagery.

The main contributions of this paper can be summarized as follows:
\begin{itemize}
\item We propose a Deformable WST Feature Bridge that adaptively aligns sonar image features with pretrained remote-sensing distributions by embedding learnable offsets into wavelet scales and orientations, effectively reducing the cross-modal degradation gap.
\item We develop a reference-free multi-frame fusion framework for sonar image enhancement, utilizing inter-frame complementary information to suppress speckle noise and enhance target contrast without external reference data.
\item We construct a carefully annotated sonar image dataset featuring three material-diverse targets (rubber tire, metal torpedo, GRP frustum) captured along circular and linear trajectories, exhibiting distinct degradation patterns (multipath scattering, acoustic absorption). Precise pose annotations eliminate the need for image registration and enable rigorous benchmarking. Experiments demonstrate superior performance of our method over state-of-the-art baselines both visually and quantitatively.
\end{itemize}

\noindent\textbf{Code and Dataset Release.} The source code and the collected sonar dataset will be publicly released upon publication.

\begin{figure*}[t]
\centering
\includegraphics[width=1\textwidth]{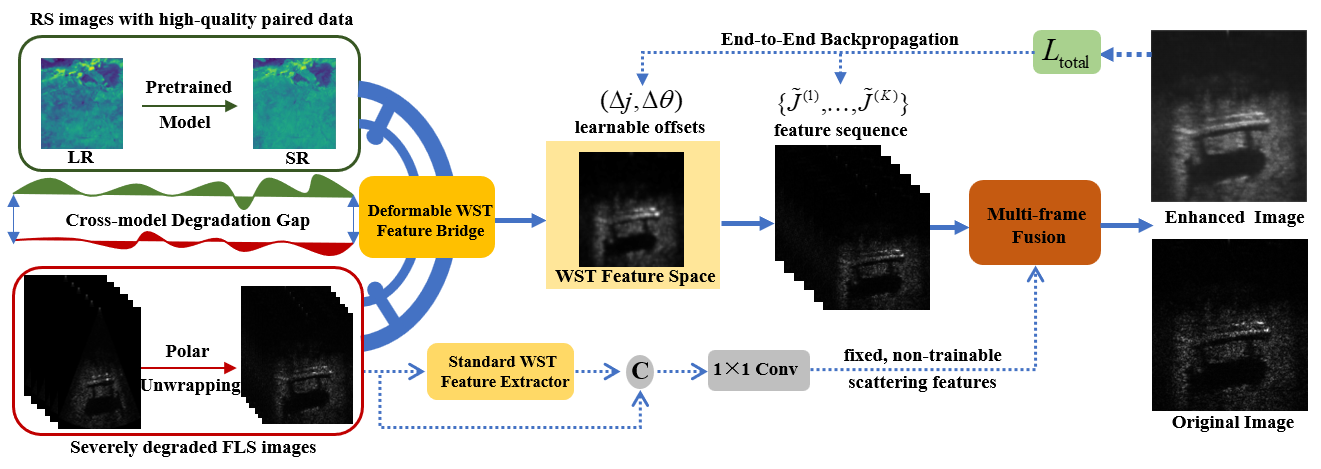}
\caption{Overview of the proposed method. 
The model combines a Deformable WST Feature Bridge—with learnable wavelet scale and orientation offsets—and a multi-frame fusion network into one end-to-end pipeline, replacing the fixed WST extractor plus simple concat-and-$1\times1$ convolution. This unified design closes the cross-modal degradation gap, reduces speckle noise, and boosts target brightness, producing images with crisper contours and richer edge detail.}
\label{fig2}
\end{figure*}

\section{Related Work}
\subsection{Quality Degradation of Forward-Looking Sonar Images and Traditional Enhancement Methods}
Forward-looking sonar images often experience quality degradation in complex underwater environments~\cite{sonar1,sonar2,sonar3,sonar4}, characterized by speckle noise, uneven illumination, and brightness artifacts due to multipath effects and acoustic absorption by target materials~\cite{sonar5,sonar6}. Among these, speckle noise is multiplicative interference from multiple acoustic paths~\cite{r5,r6}, greatly reducing target-background contrast. Traditional enhancement methods primarily aim at suppressing speckle noise\cite{r7,r8}.

These traditional methods are generally categorized into spatial-domain and transform-domain approaches. Spatial-domain methods, such as Gaussian smoothing and median filtering, reduce speckle noise but often lead to over-smoothing, causing blurred details and decreased brightness~\cite{tr3}. Transform-domain methods, including wavelet and Fourier transforms~\cite{tr1,tr2}, suppress noise and enhance details within the transformed space. However, while effectively removing high-frequency noise, these approaches frequently reduce overall image contrast and brightness.

In addition, these methods incorporate various handcrafted feature extraction techniques, such as Histogram of Oriented Gradients (HOG), Canny edge detection (Canny), Haar-like features (HAAR), Gaussian Radial Edges (GRE) and Wavelet Scattering Transform (WST)~\cite{fe1,fe2,fe3,tr1,SAR}. By extracting key structural information, these techniques transform images from the pixel domain to a more robust feature space, improving image quality under noisy conditions. We argue that such feature-space transformations effectively reduce cross-modal degradation gaps and provide a practical framework for transferring enhancement models trained in other domains to sonar image processing tasks.

\subsection{Deep-Learning-Based Sonar Image Enhancement Methods}
With the rapid advancement of deep learning, methods for enhancing and denoising forward-looking sonar images have continually evolved. Central to these methods is the challenge of generating high-quality sonar images for supervised training. For example, compressive sensing (CS) techniques and autoencoders have been applied for feature extraction in denoising, with Gaussian noise added to the original sonar images as demonstrated by~\cite{sup1,sup2}. However, both methods overlook that actual sonar image noise is multiplicative, not additive. UFIDNet~\cite{sup3} simulates this multiplicative noise by multiplying clear images with gamma-distributed noise, using logarithmic transformation to separate intensity and reflection, converting speckle noise into additive noise that is easier to remove. Furthermore,~\cite{sup4}converts optical RGBD images into 3D point clouds and uses Blender to simulate the sonar imaging process in 3D space, generating images with background noise, propagation noise, multipath effects, and azimuthal artifacts. While these methods have made progress, they rely on clear real or simulated data for supervised training, and acquiring high-quality paired data in real underwater scenarios remains challenging, limiting their generalization in practical applications.

To reduce reliance on paired data, methods combining traditional transform-domain techniques and deep learning have emerged. For instance, FDBANet~\cite{sup5} introduces a frequency-domain denoising attention module using Fast Fourier Transform (FFT) to convert sonar images to the frequency domain, filtering noise and enhancing high-frequency features. While this method works well for low-resolution images, its enhancement is limited due to the lack of a data-driven learning mechanism. reference-free methods~\cite{sr3,sr4,sr5,sr6}, however, does not require reference images and directly utilizes real low-resolution sonar images for training. The SEGSID framework~\cite{unsup1} enhances local receptive fields and extracts global semantic information, overcoming traditional blind spot network (BSN) limitations in handling speckle noise with strong spatial correlation. Yet, it uses only single-image information, limiting its ability to enhance image resolution. In contrast, multi-frame fusion methods leverage complementary information across frames, naturally removing noise and improving resolution~\cite{unsup2}. This method boosts the brightness of small target areas by fine-tuning a pre-trained remote sensing model but does not fully address the cross-modal degradation gap between multispectral remote sensing and sonar images~\cite{SAR}. In regions with complex structures or severe degradation, the enhanced image can become overly smooth, losing edge details and affecting subsequent target detection accuracy.

\section{Method}

\subsection{Framework Overview}
As shown in Figure 2, our reference-free forward-looking sonar enhancement pipeline consists of two key modules: the Deformable WST Feature Bridge and the Multi-Frame Fusion Network. The Deformable WST Feature Bridge applies learnable offsets to wavelet scales and orientations, adaptively transforming each raw sonar frame into a robust, multi-scale feature representation that aligns with pretrained remote-sensing features. These per-frame feature tensors are then stacked into a sequence and passed to the Multi-Frame Fusion Network, which comprises an encoder, a fusion block, and a decoder. By integrating complementary information across consecutive frames, the fusion network suppresses speckle noise and reconstructs a high-resolution output. All components are learned end-to-end under a reference-free objective—combining downsampling consistency with local and gradient consistency losses—and leverage accurate pose metadata to eliminate any need for explicit registration. 

\subsection{Deformable WST Feature Bridge}

To effectively bridge the cross-modal degradation gap between multispectral remote sensing and forward-looking sonar imagery, we introduce the Deformable Wavelet Scattering Transform (WST) Feature Bridge. This module adaptively transforms raw sonar images into a robust, unified feature representation tailored specifically for sonar image enhancement.

Let \(I \in \mathbb{R}^{H\times W}\) denote the input sonar image. We first obtain a set of handcrafted feature maps:
\begin{equation}
G_i(I)=\Phi_i(I),\quad i\in\{\mathrm{HOG},\mathrm{Canny},\mathrm{GRE},\mathrm{Haar},\mathrm{WST}\},
\end{equation}
where each \(\Phi_i(\cdot)\) applies the corresponding filter and \(G_i(I)\in\mathbb{R}^{H\times W}\).

Based on our ablation study, the Wavelet Scattering Transform (WST) offers an optimal trade-off between denoising and detail preservation. To further adapt WST to sonar-specific patterns, we introduce learnable scale and orientation offsets \((\Delta j, \Delta\theta)\), enabling dynamic deformation of the original Morlet wavelets:
\begin{equation}
\psi_{j,\theta}(x)\;\longmapsto\;\psi_{j+\Delta j,\,\theta+\Delta\theta}(x),\quad x\in\mathbb{R}^2,
\end{equation}
where \(j\in\{1,\dots,J\}\) and \(\theta\in\{\theta_1,\dots,\theta_K\}\), and each \(\psi\) is centered on pixel coordinate \(x\).

The first-order deformable scattering coefficients are defined by
\begin{equation}
S_1(I;j,\theta)
= \bigl\lvert\,I \ast \psi_{\,j+\Delta j,\,\theta+\Delta\theta}\bigr\rvert \ast \phi.
\end{equation}
where \(\ast\) denotes 2D convolution, \(\lvert\cdot\rvert\) the complex modulus, and \(\phi\) a fixed Gaussian low-pass filter (\(\sigma=1.0\), reflective padding) enforcing translation invariance.

Second-order deformable scattering coefficients further capture cross-scale interactions (computed only for \(j_1<j_2\) or \(\theta_1<\theta_2\) to avoid duplication):
\begin{equation}
\begin{aligned}
S_2(I;&\,j_1,\theta_1,j_2,\theta_2) \\
&= \bigl|\,
    |I \ast \psi_{j_1+\Delta j_1,\theta_1+\Delta\theta_1}| \\
&\quad\;\ast \psi_{j_2+\Delta j_2,\theta_2+\Delta\theta_2}
  \bigr|\ast\phi,
\end{aligned}
\end{equation}

Defining \(S_0=I\ast\phi\), we concatenate all orders into the unified deformable WST feature tensor:
\begin{equation}
G_{\mathrm{WST}}(I)
= \mathrm{concat}\bigl(S_0,\,S_1,\,S_2\bigr),
\end{equation}
where `concat' denotes channel-wise concatenation, yielding an output of size \(H\times W\times\bigl(1 + JK + \binom{JK}{2}\bigr)\). This tensor is then downsampled by a factor of 2 (bilinear interpolation) and channel-normalized via BatchNorm. Our implementation employs \(J=3\) scales and \(K=6\) orientations, generating \(3\!\times\!6=18\) first-order and \(\binom{18}{2}=153\) second-order channels. Crucially, all offsets \((\Delta j,\Delta\theta)\) are learned jointly with the multi-frame fusion network.

As illustrated in Fig.~\ref{fig1}(a), the original forward-looking sonar image (red) and multispectral remote sensing image (green) occupy distinct pixel domains, creating a cross-modal degradation gap. Conventional approaches attempt to narrow this gap by concatenating handcrafted feature maps with the raw image and applying a fixed \(1\times1\) convolution. In contrast, our Deformable WST Feature Bridge (Fig.~\ref{fig1}(b)) directly generates the unified, adaptively optimized feature tensor \(G_{\mathrm{WST}}(I)\), shifting data into the yellow feature space. This dynamic feature adaptation aligns with the reference-free fusion objective, substantially reducing the degradation gap and improving the transferability of pretrained remote-sensing models for sonar image enhancement tasks.

\subsection{Multi-Frame Image Fusion Network}
\begin{figure}[t]
  \centering
  \includegraphics[width=0.5\textwidth]{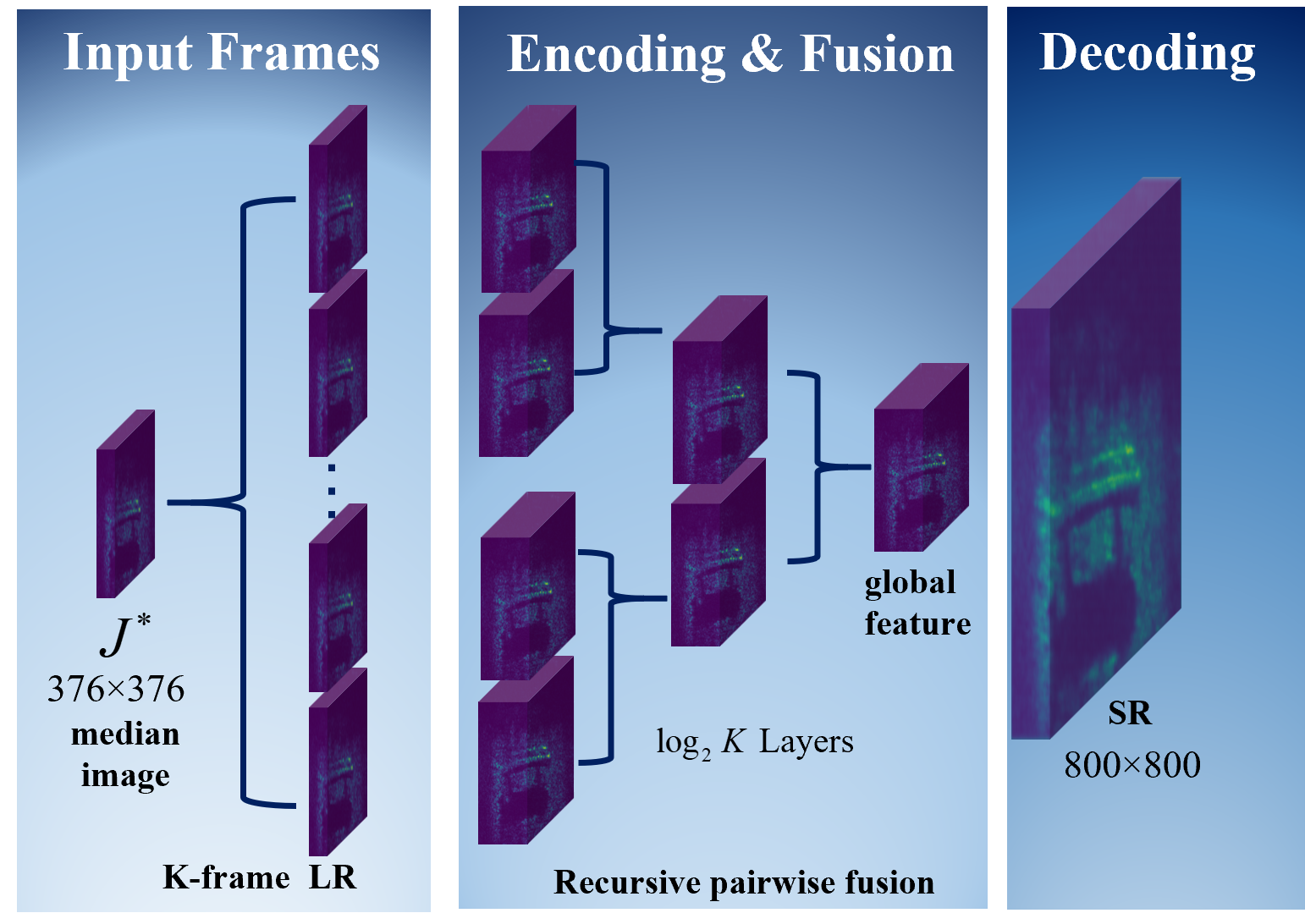}
  \caption{Illustration of the multi-frame fusion network. The network recursively fuses aligned forward-looking sonar feature tensors in a pairwise manner until all $K$ frames are aggregated. A median reference image guides reference-free enhancement through downsampling and gradient consistency losses, enabling effective speckle noise suppression and high-resolution output (from $376\times376$ to $800\times800$).}
  \label{fig3}
\end{figure}

Forward-looking sonar systems typically capture images at high frame rates, enabling complementary information across consecutive frames to naturally suppress speckle noise.  In our pipeline, each raw frame $I^{(i)}\in\mathbb{R}^{H\times W}$ is first processed by the Deformable WST Feature Bridge to produce a unified feature tensor
\[
  \tilde J^{(i)} = G_{\mathrm{WST}}\bigl(I^{(i)}\bigr),\quad i=1,\dots,K.
\]
We then form the input sequence
\[
  J = \{\tilde J^{(1)},\tilde J^{(2)},\ldots,\tilde J^{(K)}\}.
\]

The network comprises three modules: encoding $E(\cdot)$, fusion $F(\cdot)$, and decoding $D(\cdot)$:
\begin{align}
  Z &= F\bigl(E(J)\bigr), \\
  Y &= D(Z),
\end{align}
where $Y\in\mathbb{R}^{H'\times W'}$ is the super-resolved output.

For reference-free training, we adopt a downsampling mean squared error loss.  Let $H(\cdot)$ denote bicubic downsampling by factor~2, and define the pixel-wise median reference
\[
  J^*(u,v) = \operatorname{median}\bigl\{J^{(i)}(u,v)\ \big|\ i=1,\dots,K\bigr\}.
\]
The downsampling loss is
\begin{equation}
  L_{\rm down} \;=\;\bigl\lVert H(Y) - J^*\bigr\rVert_2^2.
\end{equation}

To preserve high-frequency details, we add a local consistency loss
\begin{equation}
  L_{\rm con} = \frac{1}{P}\sum_{p=1}^P\sum_{q\in\sigma(p)}
    \Bigl(\lvert Y_p - Y_q\rvert - \lvert J^*_p - J^*_q\rvert\Bigr)
\end{equation}
and a gradient consistency loss
\begin{equation}
  L_{\rm grad} = \frac{1}{P}\sum_{p=1}^P\sum_{q\in\sigma(p)}
    \Bigl(\lvert\nabla Y_p - \nabla Y_q\rvert - \lvert\nabla J^*_p - \nabla J^*_q\rvert\Bigr),
\end{equation}
where $p$ indexes local patches, $\sigma(p)$ its neighbors, and $\nabla$ the spatial gradient.

The overall objective is
\begin{equation}
  L_{\rm total} = L_{\rm down} \;+\;\lambda_{\rm con}L_{\rm con}\;+\;\lambda_{\rm grad}L_{\rm grad},
\end{equation}
with hyperparameters $\lambda_{\rm con},\lambda_{\rm grad}$ balancing each term.

\section{Experiments}
Most existing sonar image enhancement methods rely on supervised learning with simulated data, which often fails to generalize to real-world scenarios. In contrast, we compare against two recent state-of-the-art reference-free techniques—SEGSID~\cite{unsup1} and Highres-based method ~\cite{unsup2}—both of which do not require high-quality reference images. We evaluate all methods on a self-collected forward-looking sonar dataset featuring three target types. Our experiments, conducted on a 24 GB NVIDIA GeForce GTX 4090, leverage accurate pose information to avoid any additional registration before multi-frame fusion. By closing the gap between sonar and remote-sensing imagery, our Deformable WST Feature Bridge enables the direct fine-tuning of hyperspectral remote-sensing pretrained weights from~\cite{highres}, accelerating convergence and improving enhancement quality on real sonar data. Coupled with accurate pose metadata, this adaptation allows the fusion of 16 frames in just 0.3 s at inference.

\subsection{Forward-Looking Sonar Datasets}
\begin{figure}[t]
\centering
\includegraphics[width=0.5\textwidth]{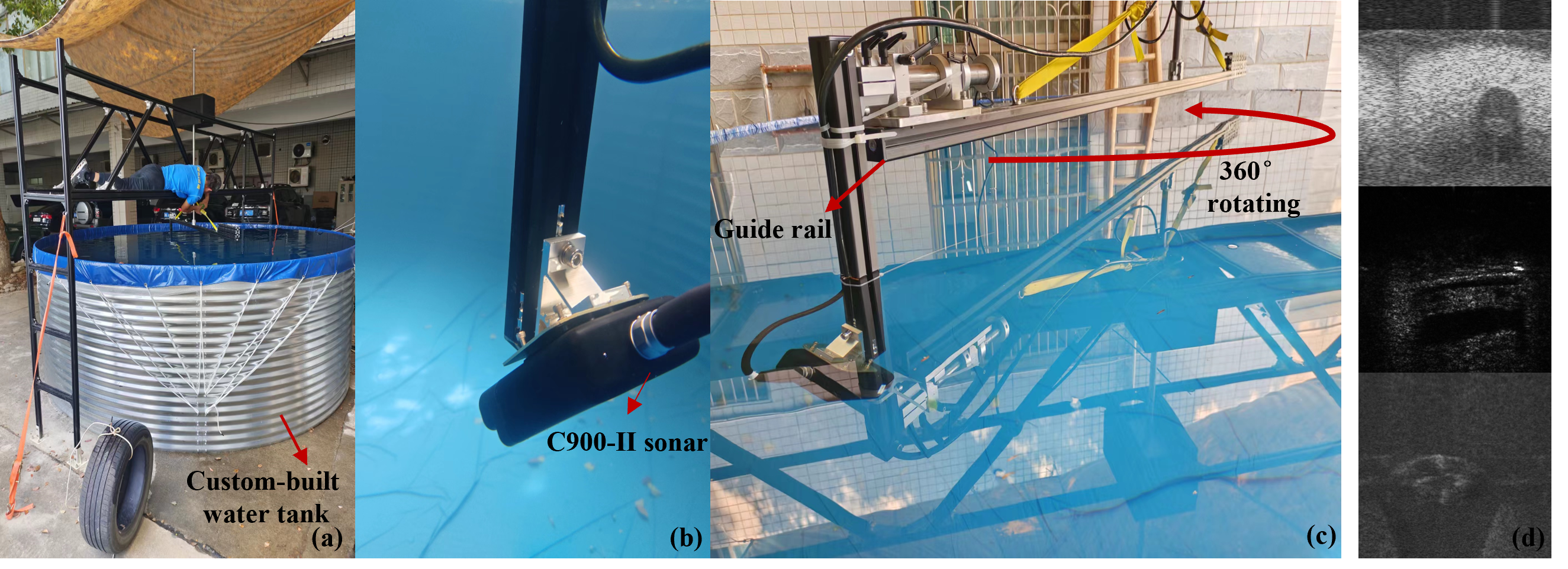}
\caption{Schematic diagram of the forward-looking sonar acquisition system and representative target images.(a) Custom-built water tank for controlled data collection; (b) C900-II sonar capable of capturing high-frame-rate sequences; (c) 360° rotating guide rail enables accurate pose estimation and multi-view imaging without additional registration; (d) three designed targets with distinct materials—rubber tire, metal torpedo model, and GRP conical frustum.}
\label{fig4}
\end{figure}
We built a standard data collection pool for forward-looking sonar (see Figure 4) using a C900-II unit operating at 2.7 MHz, with a horizontal beamwidth of 50° and a vertical beamwidth of 15°. The sonar is fixed 1.5 m from the guide rail cantilever and tilted 15° downward relative to the rotating platform. During acquisition, the system performs a 360° rotation around the target, with the guide rail rotation angle recorded continuously. Combining this rotation data with the initial sonar pose enables calculation of the sensor's pose at each instance, eliminating the need for additional image registration prior to multi-frame fusion. Each target type comprises 3,200 sonar images grouped into sequences of 16 frames each. We randomly selected 180 groups for training, and reserved 10 groups each for validation and testing. All images were converted from polar to Cartesian coordinates at a resolution of 376 × 376 pixels. Moreover, instead of training our fusion network entirely from scratch on sonar data, we initialize it with the pretrained weights provided by~\cite{unsup2}, thereby leveraging existing remote‐sensing models for faster convergence and improved generalization.

Despite being acquired in a controlled environment, our sonar images exhibit severe degradation reflective of realistic underwater scenarios, including multipath scattering, acoustic absorption, and intrinsic speckle noise. The dataset features three targets representing typical degradation conditions: \textbf{rubber tires} with severe acoustic absorption causing low contrast and blurred edges; \textbf{metal torpedo models} exhibiting complex multipath reflections and intense speckle noise; and a \textbf{glass-reinforced plastic (GRP) conical frustum} with moderate acoustic scattering characteristics commonly encountered in underwater missions. A circular trajectory ensuring full 360° coverage around each target was adopted, providing diverse angular views and accurate pose estimation without additional image registration, essential for evaluating multi-frame fusion robustness. Thus, our dataset effectively represents real-world sonar imaging challenges, validating the robustness and applicability of our enhancement method.

\subsection{Evaluation Metrics for Enhancement Performance}
In underwater environments, high‐quality reference images are scarce, making common metrics for optical image enhancement (e.g., PSNR, SSIM) impractical. Instead, we follow established reference-free sonar enhancement works~\cite{unsup1,unsup2} and employ two no‐reference metrics that capture complementary aspects of image quality:Standard deviation (STD) is used to assess image smoothness—lower STD indicates stronger noise suppression—while average gradient (AG) measures high-frequency detail, with higher AG reflecting richer detail.
\begin{equation}
STD = \sqrt{\frac{1}{N}\sum_{i=1}^{N}\left(I_i - \mu\right)^2}
\end{equation}

\begin{equation}
AG = \frac{1}{N}\sum_{i=1}^{N}\sqrt{\left(I_{x,i} - I_{x,i-1}\right)^2 + \left(I_{y,i} - I_{y,i-1}\right)^2}
\end{equation}

where ${I_i}$ denotes the grayscale value of the $i^{\text{th}}$ pixel, ${I_{x,i}}$ and ${I_{y,i}}$ represent the horizontal and vertical gradients, respectively, ${\mu}$ is the mean value of the image, and ${N}$ is the total number of pixels.

\subsection{Quantitative and Visual Evaluation of Enhancement Results}

Table~\ref{table1} presents a quantitative comparison across three representative target types: torpedo model, tire, and conical frustum. Our method consistently achieves the best results in both standard deviation (STD) and average gradient (AG), demonstrating strong capabilities in suppressing noise and preserving structural details.

For the torpedo model, the STD is reduced from 17.02 to 13.14, surpassing SEGSID (21.59) and HighRes-based (14.58) methods. AG improves from 4.93 to 5.92, indicating sharper edges and clearer target contours.

In the tire category—challenging due to strong acoustic absorption—our method delivers the most substantial improvement, reducing STD from 58.79 to 37.68 and boosting AG from 8.09 to 14.71. This highlights its robustness in handling severe degradation while maintaining fine details.

For the conical frustum, our method again outperforms all baselines, achieving the lowest STD (23.24) and highest AG (13.66), confirming its adaptability to varying materials and geometries.

Overall, the proposed method demonstrates consistently superior performance across all scenarios. By combining feature space transformation with multi-frame fusion, it effectively balances speckle noise suppression and detail preservation—crucial for accurate underwater target detection in forward-looking sonar imagery.

\begin{figure}[!t]
\centering
\includegraphics[width=0.48\textwidth]{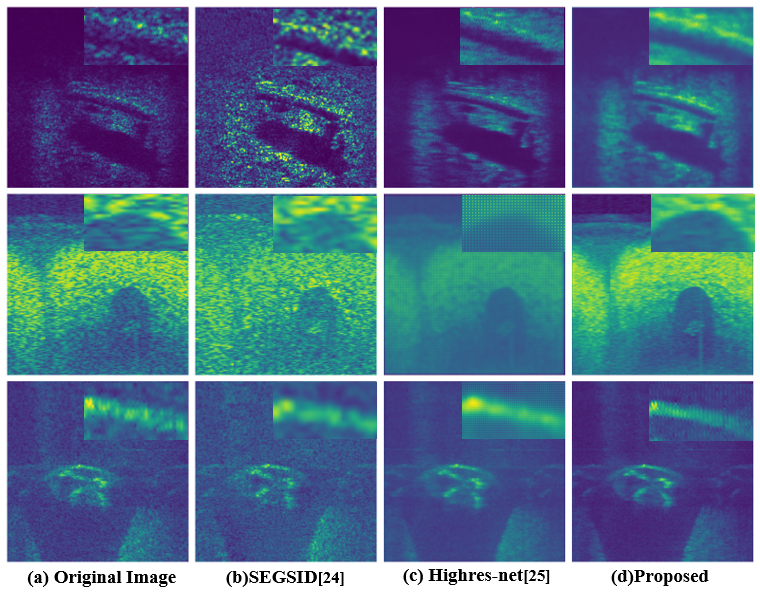}
\caption{Visual comparison of enhancement performance across three target categories.
Each row shows enlarged target regions (torpedo model, tire, and conical frustum) for detailed comparison. Existing methods either amplify noise or over-smooth targets, leading to blurred edges and poor contrast. In contrast, our method enhances brightness, preserves clear contours, and suppresses background noise effectively.}
\label{fig5}
\end{figure}

These quantitative improvements are further supported by visual comparisons. Figure~\ref{fig5} presents enhancement results for the same three targets, with each row showing enlarged views of the target regions for detailed inspection. Existing methods either amplify background noise or blur target contours, leading to poor contrast and reduced structural clarity. In contrast, our method enhances target intensity and edge sharpness while maintaining a clean background, closely aligning with the improvements observed in Table~\ref{table1}.

Furthermore, Figure~\ref{fig8} illustrates enhancement results under different sonar viewpoints. The original degraded images exhibit strong speckle noise and weak contrast. After applying feature space transformation and multi-frame fusion, our method produces significantly brighter target regions and clearer boundaries, demonstrating robustness to viewpoint variation and structural deformation.

In summary, these quantitative and visual results together validate the robustness and effectiveness of our proposed method. By integrating feature space transformation with multi-frame fusion, the approach consistently achieves superior performance across diverse targets and degradation conditions. It excels in suppressing speckle noise, preserving fine structural details, and maintaining stable enhancement across varying viewpoints—highlighting its strong generalizability and practical value for real-world forward-looking sonar image enhancement.

\begin{table}[!t]
\centering
\caption{Enhancement performance comparison for three different target types.}
\scriptsize
\renewcommand\arraystretch{1.2}
\setlength{\heavyrulewidth}{0.08em}
\setlength{\lightrulewidth}{0.05em}
\setlength{\cmidrulewidth}{0.03em}

\begin{minipage}{0.48\textwidth}
\centering
\textbf{(a) Torpedo model } \\
\vspace{0.3em}
\begin{tabular}{lcccc}
\toprule
Fusion Method & Original & SEGSID [24] & Highres [25] & Proposed \\
\midrule
STD $\downarrow$ & 17.02 & 21.59 & 14.58 & \textbf{13.14} \\
AG $\uparrow$    & 4.93  & 5.06  & 3.70  & \textbf{5.92} \\
\bottomrule
\end{tabular}
\end{minipage}
\vspace{1em}

\begin{minipage}{0.48\textwidth}
\centering
\textbf{(b) Tire } \\
\vspace{0.3em}
\begin{tabular}{lcccc}
\toprule
Fusion Method & Original & SEGSID [24] & Highres [25] & Proposed \\
\midrule
STD $\downarrow$ & 58.79 & 45.96 & 48.80 & \textbf{37.68} \\
AG $\uparrow$    & 8.09  & 9.61  & 10.62 & \textbf{14.71} \\
\bottomrule
\end{tabular}
\end{minipage}
\vspace{1em}

\begin{minipage}{0.48\textwidth}
\centering
\textbf{(c) Conical frustum} \\
\vspace{0.3em}
\begin{tabular}{lcccc}
\toprule
Fusion Method & Original & SEGSID [24] & Highres [25] & Proposed \\
\midrule
STD $\downarrow$ & 30.41 & 30.42 & 28.83 & \textbf{23.24} \\
AG $\uparrow$    & 9.99  & 7.99  & 10.32 & \textbf{13.66} \\
\bottomrule
\end{tabular}
\end{minipage}

\label{table1}
\end{table}

\begin{table}[!t]
\centering
\caption{Enhancement performance of different feature operators for the torpedo model target.}
\scriptsize
\renewcommand\arraystretch{1.2}
\setlength{\heavyrulewidth}{0.08em}
\setlength{\lightrulewidth}{0.05em}
\setlength{\cmidrulewidth}{0.03em}

\begin{minipage}{0.9\linewidth} 
\centering
\begin{tabular}{p{3.2cm}cc} 
\toprule
\textbf{Input} & \textbf{STD $\downarrow$} & \textbf{AG $\uparrow$} \\
\midrule
FLR                  & 17.016 & 4.932 \\
FLR + HOG            & 27.338 & 4.809 \\
FLR + Canny          & \textbf{8.978}  & 1.552 \\
FLR + GRE            & 16.277 & 3.131 \\
FLR + HAAR           & 10.348 & 1.683 \\
FLR + WST            & 13.143 & \textbf{5.918} \\
\bottomrule
\end{tabular}
\end{minipage}

\label{table2}
\end{table}

\begin{figure}[t]
  \centering
  \includegraphics[width=0.5\textwidth]{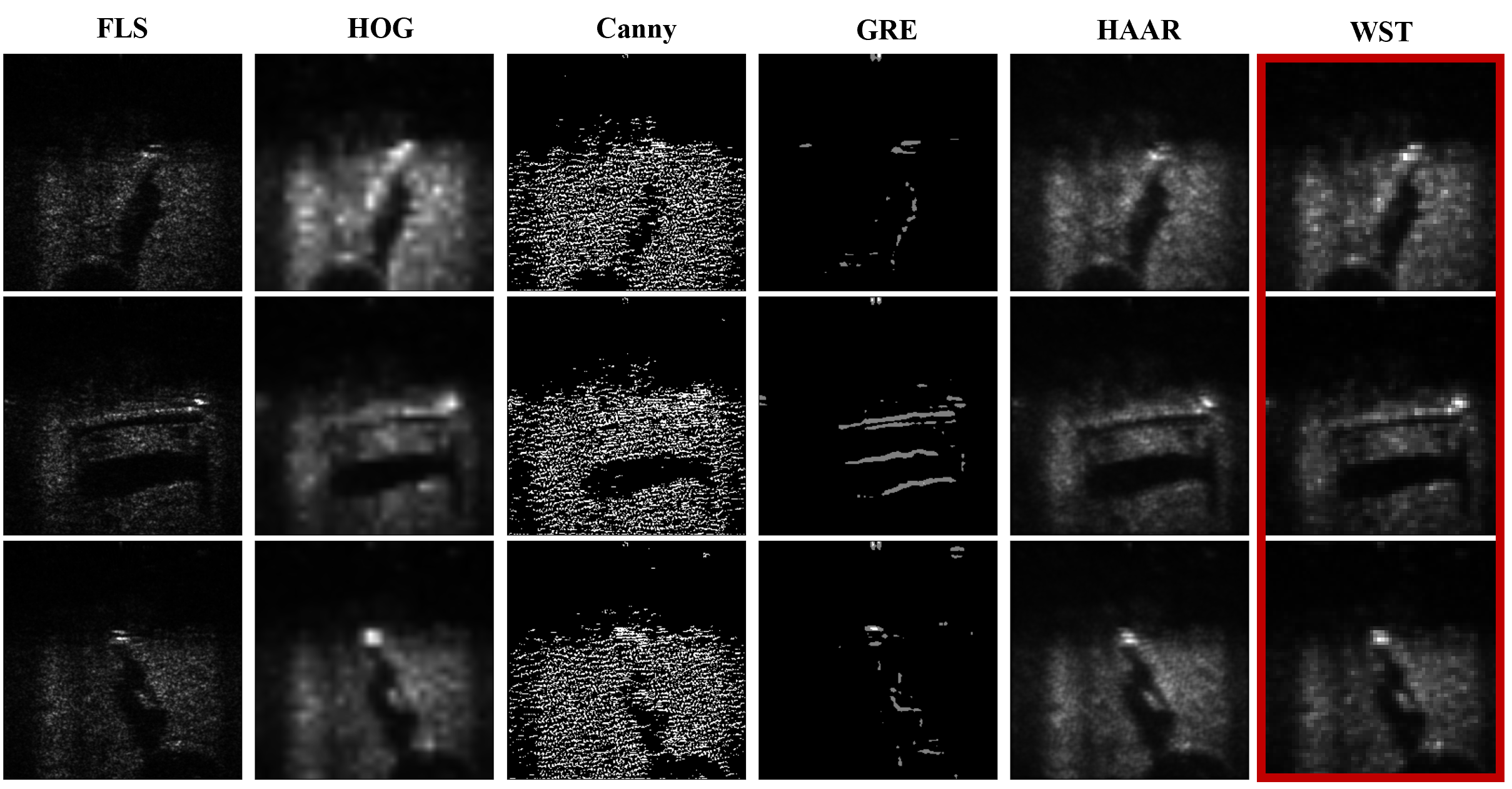}
  \caption{Visualization of handcrafted features. WST is selected for its balanced performance in denoising and detail retention.}
  \label{fig6}
\end{figure}

\begin{figure}[t]
\centering
\includegraphics[width=0.48\textwidth]{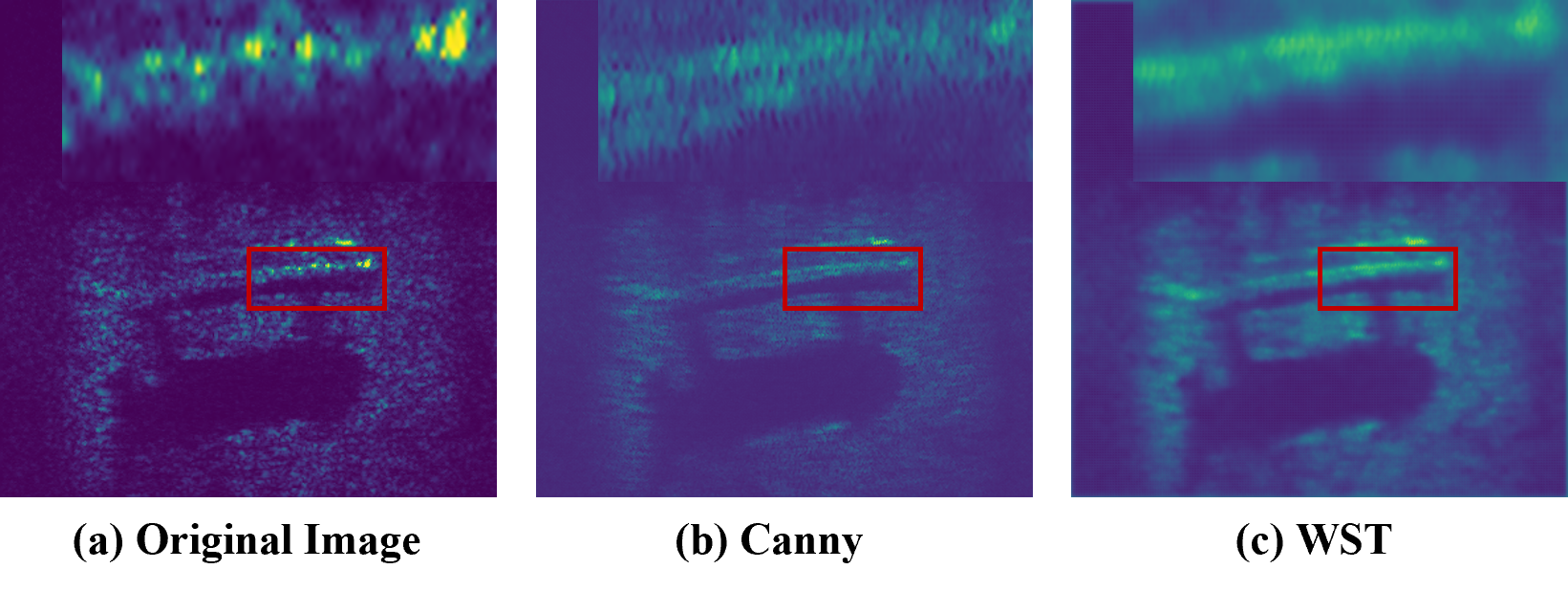}
\caption{Comparison of enhancement results using Canny and WST transformations.
WST preserves high-frequency details with stronger target contrast and clearer boundaries, while Canny produces oversmoothed results with blurred edges, limiting detection accuracy.}
\label{fig7}
\end{figure}

\subsection{Ablation Study}
Figure~\ref{fig6} shows the handcrafted feature operators (HOG, Canny, GRE, HAAR, WST). We perform ablation experiments within our multi-frame reference-free enhancement framework to demonstrate that WST achieves the best balance between noise suppression and detail preservation, justifying its selection as the primary feature operator.

\begin{figure*}[!t]
\centering
\includegraphics[width=1\textwidth]{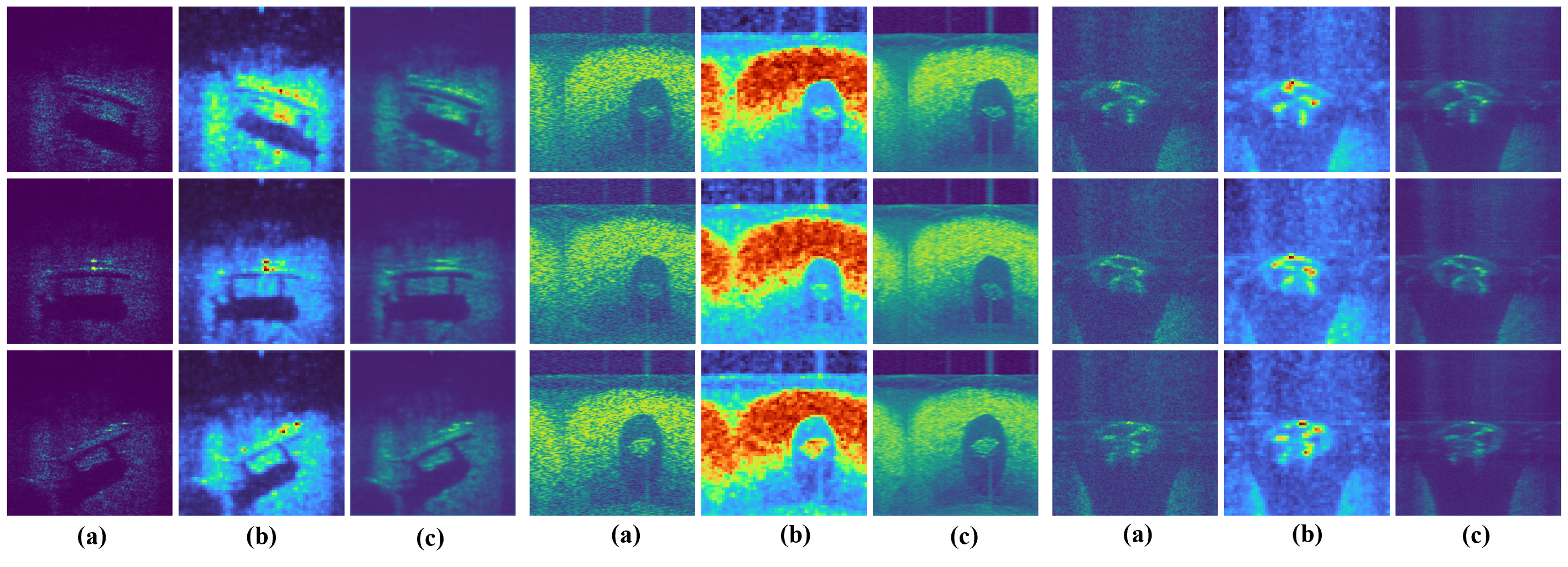}
\caption{Enhancement results across varying sonar viewpoints.
(a) Original images exhibit heavy speckle noise and low contrast across perspectives. (b) Feature space transformation reduces cross-modal discrepancies and reinforces structural consistency. (c) The proposed method enhances brightness and edge clarity, achieving robust performance under viewpoint changes.}
\label{fig8}
\end{figure*}
Table~\ref{table2} reports quantitative results in terms of standard deviation (STD, lower is better for noise suppression) and average gradient (AG, higher is better for detail preservation). While Canny achieves the lowest STD—indicating strong noise removal—its AG is drastically lower than other methods, reflecting significant loss of edge information. In contrast, WST attains a near-optimal STD reduction and the highest AG, striking the best balance between denoising and detail retention. Figure~\ref{fig7} illustrates this: WST enhances intensity and edge clarity, whereas Canny’s output, despite being smoother, suffers from blurred boundaries. Hence, although Canny suppresses noise more aggressively, WST’s balanced performance makes it the preferred choice for our feature transformation module.

\section{Conclusion}
We propose a reference-free framework for forward‐looking sonar enhancement that integrates a Deformable WST Feature Bridge with an end‐to‐end multi‐frame fusion network. The Deformable WST bridge uses learnable wavelet scale and orientation offsets to align sonar features with pretrained remote‐sensing distributions, closing the cross‐modal gap. The fusion network then aggregates complementary frames to suppress speckle noise and boost target brightness—without any external ground truth. On three self‐collected datasets (rubber tire, metal torpedo, GRP frustum), our method outperforms all reference‐free baselines in STD, AG, and visual quality. By narrowing the degradation gap, it also allows direct fine‐tuning of remote‐sensing pretrained weights for faster convergence and stronger generalization. Achieving 16‐frame processing in just 0.3 s on a single GPU, it is well suited for real‐time underwater deployment. Future work will validate the approach in field environments and embed it within autonomous underwater vehicles.

\section{Acknowledgment}
This work was supported by the National Natural Science Foundation of China (No. 62401601).
\bibliographystyle{IEEEtran}
\bibliography{reference}

\end{document}